\documentclass{article}

\usepackage[final]{corl_2017} 

\usepackage{graphicx}
\usepackage{caption} 
\usepackage{wrapfig}
\title{Transferring End-to-End Visuomotor Control from Simulation to Real World for a Multi-Stage Task}

\author{
  Stephen James\\
  Dyson Robotics Lab\\
  Imperial College London\\
  \texttt{slj12@ic.ac.uk}\\
  \And
  Andrew J. Davison\\
  Dyson Robotics Lab\\
  Imperial College London\\
  \texttt{a.davison@ic.ac.uk}\\
  \And
  Edward Johns\\
  Dyson Robotics Lab\\
  Imperial College London\\
  \texttt{e.johns@ic.ac.uk}\\
}

\begin{document}
\maketitle


\begin{abstract}
End-to-end control for robot manipulation and grasping is emerging as an attractive alternative to traditional pipelined approaches. However, end-to-end methods tend to either be slow to train, exhibit little or no generalisability, or lack the ability to accomplish long-horizon or multi-stage tasks. In this paper, we show how two simple techniques can lead to end-to-end (image to velocity) execution of a multi-stage task, which is analogous to a simple tidying routine, without having seen a single real image. This involves locating, reaching for, and grasping a cube, then locating a basket and dropping the cube inside. To achieve this, robot trajectories are computed in a simulator, to collect a series of control velocities which accomplish the task. Then, a CNN is trained to map observed images to velocities, using domain randomisation to enable generalisation to real world images. Results show that we are able to successfully accomplish the task in the real world with the ability to generalise to novel environments, including those with dynamic lighting conditions, distractor objects, and moving objects, including the basket itself. We believe our approach to be simple, highly scalable, and capable of learning long-horizon tasks that have until now not been shown with the state-of-the-art in end-to-end robot control.
\end{abstract}



\section{Introduction}
\label{sec:introduction}

An emerging trend for robot manipulation and grasping is to learn controllers directly from raw sensor data in an end-to-end manner. This is an alternative to traditional pipelined approaches which often suffer from propagation of errors between each stage of the pipeline. End-to-end approaches have had success both in the real world~\citep{levine2016end, montgomery2016guided, montgomery2016reset} and in simulated worlds~\citep{popov2017data,james20163d, zhang2015towards, higgins2017darla}. Learning end-to-end controllers in simulation is an attractive alternative to using physical robots due to the prospect of scalable, rapid, and low-cost data collection. However, these simulation approaches are of little benefit if we are unable to transfer the knowledge to the real world. What we strive towards are robust, end-to-end controllers that are trained in simulation, and can run in the real world without having seen a single real world image. 

\begin{figure}[]
\centering
\includegraphics[width=1.0\linewidth]{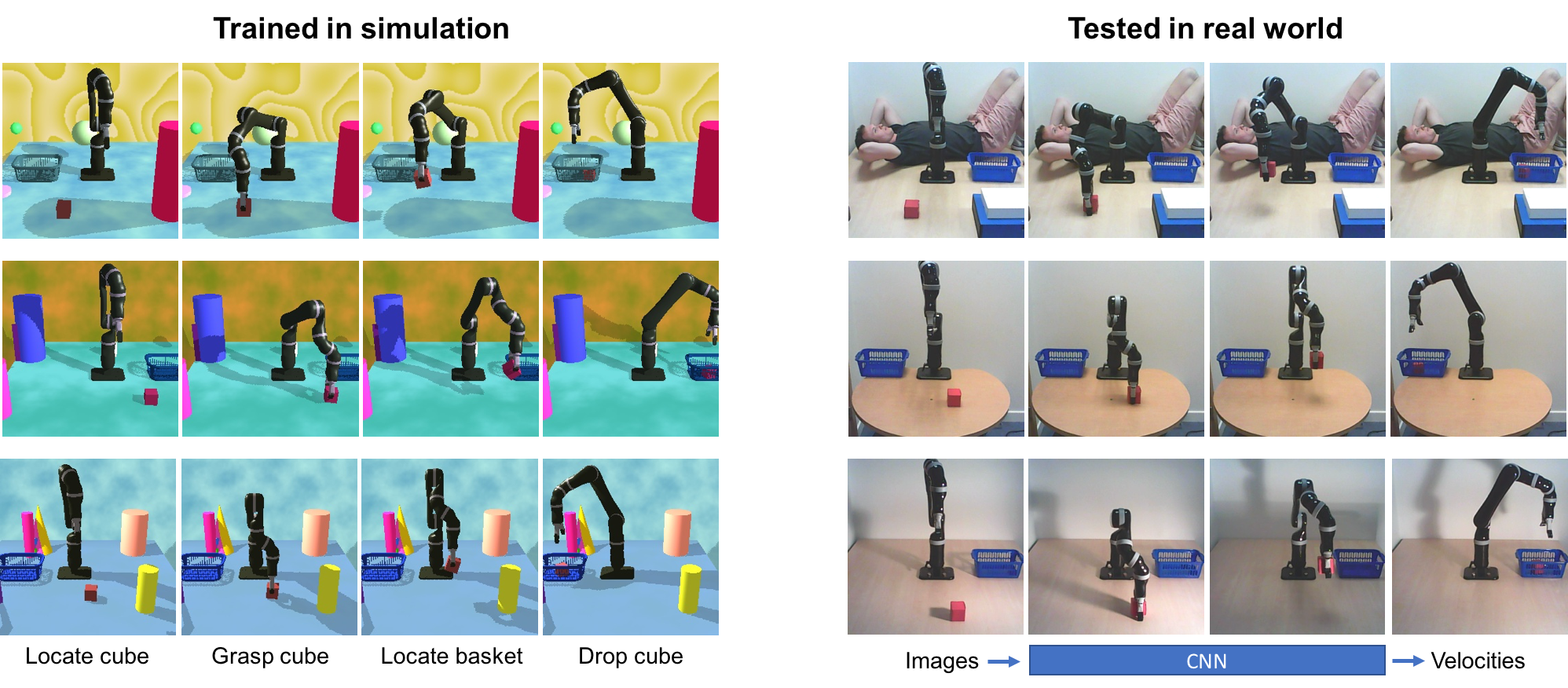}
\caption{Our approach uses simulation to collect a series of control velocities to solve a multi-stage task. This is used to train a reactive neural network controller which continuously accepts images and joint angles, and outputs motor velocities. By using domain randomisation, the controller is able to run in the real world without having seen a single real image.}
\label{fig:front}
\end{figure}

In this paper, we accomplish the goal of transferring end-to-end controllers to the real world and demonstrate this by learning a long-horizon multi-stage task that is analogous to a simple tidying task, and involves locating a cube, reaching, grasping, and locating a basket to drop the cube in. This is accomplished by using demonstrations of linear paths constructed via inverse kinematics (IK) in the Cartesian space, to construct a dataset that can then be used to train a reactive neural network controller which continuously accepts images along with joint angles, and outputs motor velocities. We also show that task performances improves with the addition of auxiliary outputs (inspired by~\citep{jaderberg2016reinforcement,dilokthanakul2017feature}) for positions of both the cube and gripper. Transfer is made possible by simply using domain randomisation~\citep{sadeghi2016cad, tobin2017domain}, such that through a large amount of variability in the appearance of the world, the model is able to generalise to real world environments. Figure \ref{fig:front} summarises the approach, whilst our video demonstrates the success of the final controller in the real world \footnote{\label{foot:video}Video: \url{https://youtu.be/X3SD56hporc}}.

Our final model is not only able to run on the real world, but can accomplish the task with variations in the position of the cube, basket, camera, and initial joint angles. Moreover, the model shows robustness to distractors, lighting conditions, changes in the scene, and moving objects (including people). 


\section{Related Work}
\label{sec:relatedWork}

End-to-end methods for control are often trained using Reinforcement Learning (RL). In the past few years, classic RL algorithms have been fused with function approximators, such as neural networks, to spawn the domain of deep reinforcement learning, which is capable of playing games such as Go~\citep{silver2016mastering} and Atari~\citep{mnih2015human} to a super-human level. There have been advances in applying these techniques to both simulated robotic platforms~\citep{popov2017data, james20163d}, and real-world platforms~\citep{gu2016deep}, but fundamental challenges remain, such as sample inefficiency, slow convergence, and the algorithm's sensitivity to hyperparameters. There have been attempts at transferring trained polices from the real world following training in simulation, but these attempts have either failed~\citep{james20163d}, or required additional training in the real world~\citep{rusu2016sim}. Although applying RL can work well for computer games and simulations, the same cannot be said as confidently for real-world robots, where the ability to explore sufficiently can become intractable as the complexity of the task increases. To counter this, imitation learning can be used by providing demonstrations in first~\citep{duan2017one, lee2015learning, hausman2017multi} or third~\citep{stadie2017third} person perspective. Our method uses the full state of the simulation to effectively produce first-person demonstrations for supervision without the need for exploration.

A different approach for end-to-end control is guided policy search (GPS)~\citep{levine2013guided, levine2014learning}, which has achieved great success particularly in robot manipulation~\citep{levine2016end, montgomery2016guided, montgomery2016reset}. Unlike most RL methods, these approaches can be trained on real world robotic platforms, but therefore have relied on human involvement which limits their scalability. To overcome human involvement, GPS has been used with domain adaptation of both simulated and real world images to map to a common feature space for pre-training~\citep{tzeng2016adapting}; however, this still requires further training in the real world. In comparison, our method has never seen a real world image before, and learns a controller purely in simulation.

One approach to scale up available training data is to continuously collect data over long periods of time using one~\citep{pinto2016supersizing} or multiple robots~\citep{levine2016learning, finn2016deep}. This was the case for~\citep{levine2016learning}, where 14 robots were run for a period of 2 months and collected $800,000$ grasp attempts by randomly performing grasps. A similar approach was used to learn a predictive model in order to push objects to desired locations~\citep{finn2016deep}. Although the results are impressive, there is some doubt in scalability with the high purchase cost of robots, in addition to the question of how you would get data for more complex and long-horizon tasks. Moreover, these solutions typically cannot generalise to new environments without also training them in that same environment.

There also exist a number of works that do not specifically learn end-to-end control, but do in fact use simulation to learn behaviours with the intention to use the learned controller in the real world. Such works include~\citep{johns2016deep}, which uses depth images from simulated 3D objects to train a CNN to predict a score for every possible grasp pose. This can then be used to locate a grasp point on novel objects in the real world. Another example is~\citep{christiano2016transfer}, where deep inverse models are learned within simulation to perform a back-and-forth swing of a robot arm using position control. For each time step during testing, they query the simulated control policy to decide on suitable actions to take in the real world.

Transfer learning is concerned with transferring knowledge between different tasks or scenarios. Works such as \citep{devin2017learning, gupta2017learning} show skill transfer within simulation, whereas we concentrate on simulation-to-real transfer. In~\citep{sadeghi2016cad}, the focus is on learning collision-free flight in simulated indoor environments using realistic textures sampled from a dataset. They show that the trained policy can then be directly applied to the real world. Their task differs to ours in that they do not require hand-eye coordination, and do not need to deal with a structured multi-stage task. Moreover, rather than sampling from a dataset of images, ours are procedurally generated, which allows for much more diversity. During development of our work, a related paper emerged which uses the domain randomisation~\citep{tobin2017domain} method in a similar manner to us, except that the focus is on pose estimation rather than end-to-end control. We operate at the lower level of velocity control, to accomplish a multi-stage task which requires not only pose estimation, but also target reaching and grasping. In addition, we show that our learned controller can work in a series of stress tests, including scenes with dramatic illumination changes and moving distractor objects.


\vspace{-1.5 mm}

\section{Approach}
\label{sec:approach}

\vspace{-2.5 mm}

\begin{figure}[]
\centering
\includegraphics[width=1.0\linewidth]{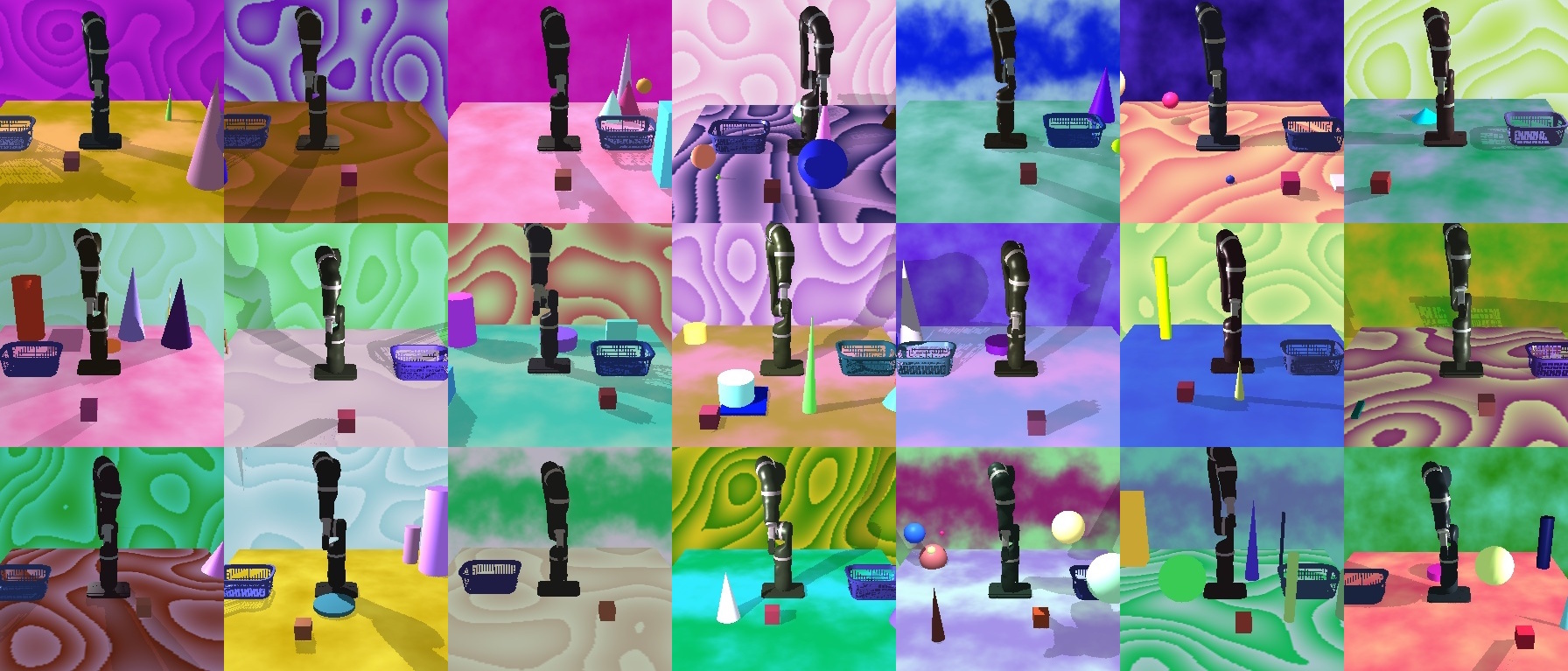}
\caption{A collection of generated environments as part of our domain randomisation method.}
\label{fig:data_gen}
\vspace{-1.9mm}
\end{figure}

Our aim is to create an end-to-end reactive controller that does not require real-world data to train, and is able to learn complex behaviours in a short period of time. To achieve this, we generate a large number of trajectories in simulation, together with corresponding image observations, and then train a controller to map observed images to motor velocities, which are effected through a PID controller. Through the use of domain randomisation during the data generation phase, we are able to run the controller in the real world without having seen a single real image. We now describe in detail the dataset generation and training method. 

\subsection{Data Collection}

The success of this work comes down to the way in which we generate the training data (Figure \ref{fig:data_gen}). Our approach uses a series of linear paths constructed in the Cartesian space via inverse kinematics (IK) in order to construct the task sequence. At each simulation step, we record motor velocities, joint angles, gripper actions (open or close command), cube position, gripper position, and camera images. We split the task into 5 stages, and henceforth, we refer to the sequence of stages as an \textit{episode}. At the start of each episode, the cube and basket is placed randomly within an area that is shown in Figure \ref{fig:vrep_cam_variations}. We use the V-REP~\citep{vrep} simulator during all data collection.

In the first stage of the task, the arm is reset to an initial configuration. We place a waypoint above the cube and plan a linear path which we then translate to motor velocities to execute. Once this waypoint has been reached, we execute the second stage, where by a closing action on the gripper is performed. The third stage sets a waypoint a few inches above the cube, and we plan and execute a linear path in order to lift the cube upwards. The fourth stage places a waypoint above the basket, where we plan and execute a final linear path to take the grasped cube above the basket. Finally, the fifth stage simply performs a command that opens the gripper to allow the cube to fall into the basket. A check is then carried out to ensure that the location of the cube is within the basket; if this is the case, then we save the episode. We do not consider obstacle avoidance in this task, and so episodes that cannot find a linear set of paths due to obstacles are thrown away. The data generation method can be run on multiple threads, which allows enough data for a successful model to be collected within a matter of hours, and increasing the number of threads would further reduce this time further.

Using this approach, it is already possible to train a suitable neural network to learn visuomotor control of the arm and perform the task to succeed 100\% of the time when tested in simulation. However, we are concerned with applying this knowledge to the real world so that it is able to perform equally as well as it did in simulation. By using domain randomisation, we are able to overcome the reality-gap that is present when trying to transfer from the synthetic domain to the real world domain. For each episode, we list the environment characteristics that are varied, followed by an explanation of why we vary them:

\begin{figure}[]
\centering
\includegraphics[width=0.6\linewidth]{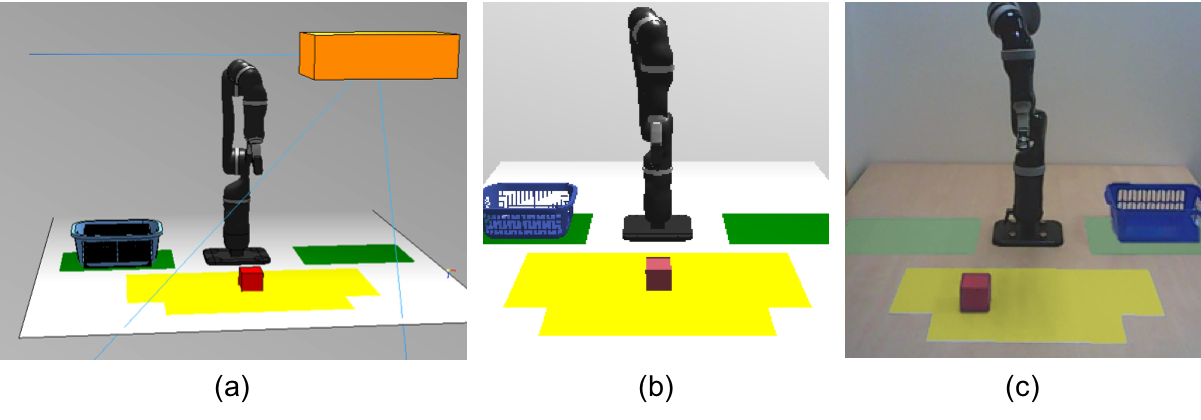}
\caption{Variations in the positioning of the cube, basket, and camera, during both training and testing. In all three images, the yellow area represents the possible locations of the cube, and the green areas represent the possible locations of the basket. (a) shows variations of the camera pose, illustrated by the orange cuboid. (b) shows an example view from the camera in simulation. (c) shows an example view from the camera in the real world. Note that the yellow, green, and orange shapes are purely for visualisation here, and are not seen during training or testing.}
\label{fig:vrep_cam_variations}
\end{figure}

\begin{itemize}

\item The colour of the cube, basket, and arm components are sampled from a normal distribution, with the mean set as close to the estimated real world equivalent; though these could also be sampled uniformly. 

\item The position of the camera, light source (shadows), basket, and cube are sampled uniformly. Orientations are kept constant.

\item The height of the arm base from the table is sampled uniformly from a small range.

\item The starting joint angles are sampled from a normal distribution with the mean set to the configuration in Figure \ref{fig:vrep_cam_variations}.

\item We make use of Perlin noise~\citep{perlin2002improving} composed with functions (such as sine waves) to generate textures, which are then applied to the table and background of the scene.
 
\item We add random primitive shapes as distractors, with random colours, positions, and sizes sampled from a uniform distribution. 
\end{itemize}

A selection of these generated environments can be seen in Figure \ref{fig:data_gen}. We chose to vary the colours and positions of the objects in the scene as part of a basic domain randomisation process. Slightly less obvious is varying the height of the arm base; this is to avoid the network learning a controller for a set height, that we cannot guarantee is the true height in the real world. In order to successfully run these trained models in the real world, we also must account for non-visual issues, such as the error in the starting position of the joints when run on the real world. Although we could send the real world arm to the same starting position as the synthetic arm, in practice the joint angles will be slightly off, and this could be enough to put the arm in configurations that are unfamiliar and lead to compounding errors along its trajectory. It is therefore important that the position of the robot at the start of the task is perturbed during data generation. This leads to controllers that are robust to compounding errors during execution, since a small mistake on the part of the learned controller would otherwise put it into states that are outside the distribution of the training data. This method can also be applied to the generated waypoints, but in practice this was not needed. We use procedural textures rather than plain uniform textures, to achieve sufficient diversity which encompasses background textures of the real world.

\subsection{Network Architecture}

\begin{figure}[]
\centering
\includegraphics[width=1.0\linewidth]{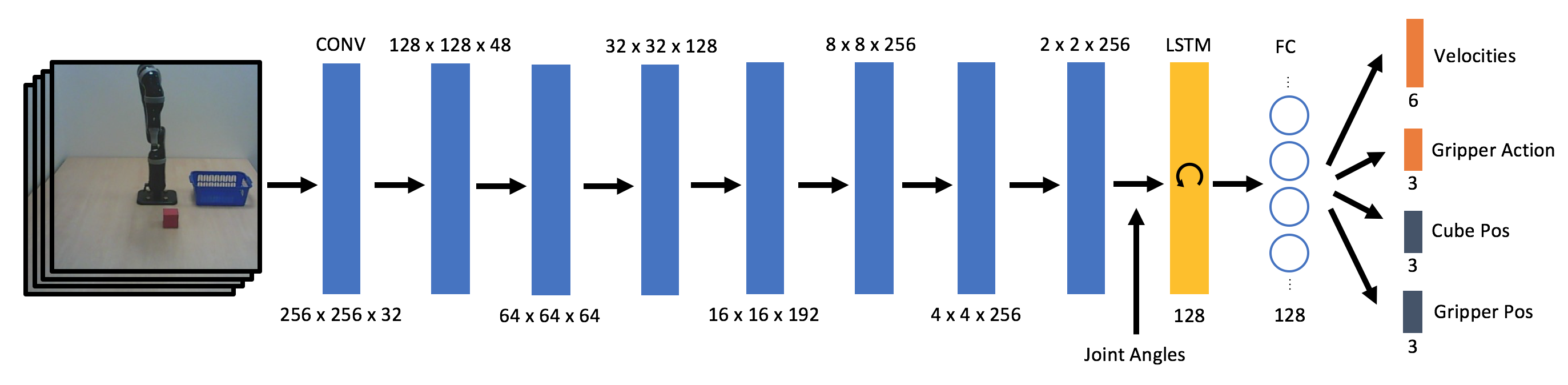}
\caption{Our network architecture continuously maps sequences of the past 4 images and joint angles to motor velocities and gripper actions, in addition to 2 axillary outputs: the 3D cube position and 3D gripper position. These axillary outputs are not used during testing, but are instead present to help the network learn informative features.}
\label{fig:network}
\end{figure}

The network, summarised in Figure \ref{fig:network}, consists of 8 convolutional layers each with a kernel size of $3 \times 3$, excluding the last, which has a size of $2 \times 2$. Dimensionality reduction is performed at each convolutional layer by using a stride of 2. Following the convolutional layers, the output is concatenated with the joint angles and then fed into an LSTM module (we discuss the importance of this in the results). Finally, the data goes through a fully-connected layer of 128 neurons before heading to the output layer. We chose to output velocities, effected via a PID controller, rather than to output torques directly, due to the difficulty in transferring complex dynamics from simulation.

The network outputs 6 motor velocities, 3 gripper actions, and 2 auxiliary outputs: cube position and gripper position. We treat the 3 gripper actions as a classification problem, where the outputs are $\{open, close, no$-$op\}$. During testing, the auxiliary outputs are not used at any point, but are present to aid learning and conveniently help debug the network. By rendering a small marker at the same positions as the auxiliary outputs, we are able to observe where the controller estimates the cube is, in addition to where it estimates the gripper is, which can be helpful during debugging.

Our loss function $\mathcal{L}_{Total}$ is a combination of the mean squared error of the velocities ($V$) and gripper actions ($G$), together with the gripper position ($GP$) auxiliary and cube position ($CP$) auxiliary, giving

\vspace{-5pt}

\begin{equation}
\mathcal{L}_{Total} = \mathcal{L}_{V} + \mathcal{L}_{G} + \mathcal{L}_{GP} + \mathcal{L}_{CP}.
\end{equation}

We found that simply weighting the loss terms equally resulted in effective and stable training. The model was trained with the Adam optimiser~\citep{kingma2014adam} with a learning rate of $10^{-4}$.


\section{Experiments}
\label{sec:results}

In this section we present results from a series of experiments, not only to show the success of running the trained models in different real world settings, but also to show what aspects of the domain randomisation are most important for a successful transfer. Figure \ref{fig:basket_move_timeline} shows an example of the controller's ability to generalise to new environments in the real world, although many more are shown in the video\footnote{\label{foot:video}Video: \url{https://youtu.be/X3SD56hporc}}. We focused our experiments to answer the following questions:

\begin{figure}[]
\centering
\includegraphics[width=1.0\linewidth]{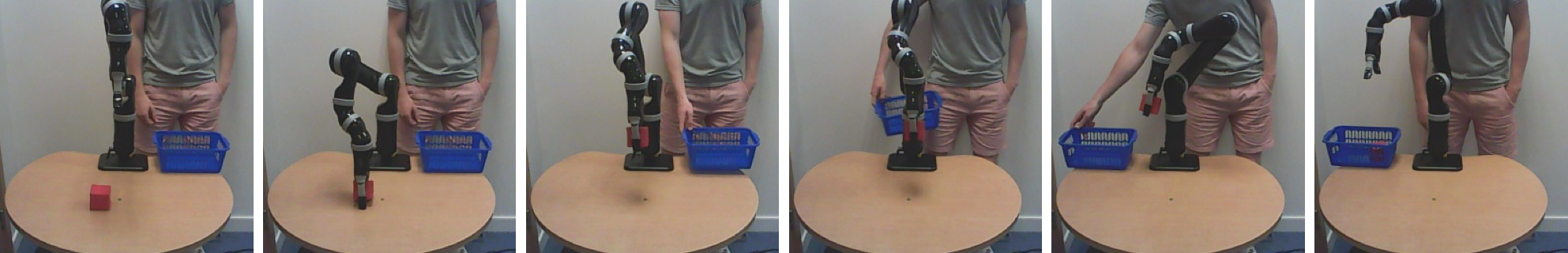}
\caption{A sequence of images showing the network's ability to generalise. Whilst a person is standing in the view of the camera, the robot is able to grasp the cube and proceed to the basket. As the arm progresses towards the basket, the person moves the basket to the other side of the arm. Despite this disruption, the network alters its course and proceeds to the new location. During training, the network had never seen humans, moving baskets, or this particular table.}
\label{fig:basket_move_timeline}
\end{figure}

\begin{enumerate}
\item How does performance vary as we alter the dataset size?
\item How robust is the trained controller to new environments?
\item What is most important to randomise during domain randomisation?
\item Does the addition of auxiliary outputs improve performance?
\item Does the addition of joint angles as input to the network improve performance?
\end{enumerate}

\begin{wrapfigure}[13]{r}{5.2cm}
\centering
\includegraphics[width=0.9\linewidth]{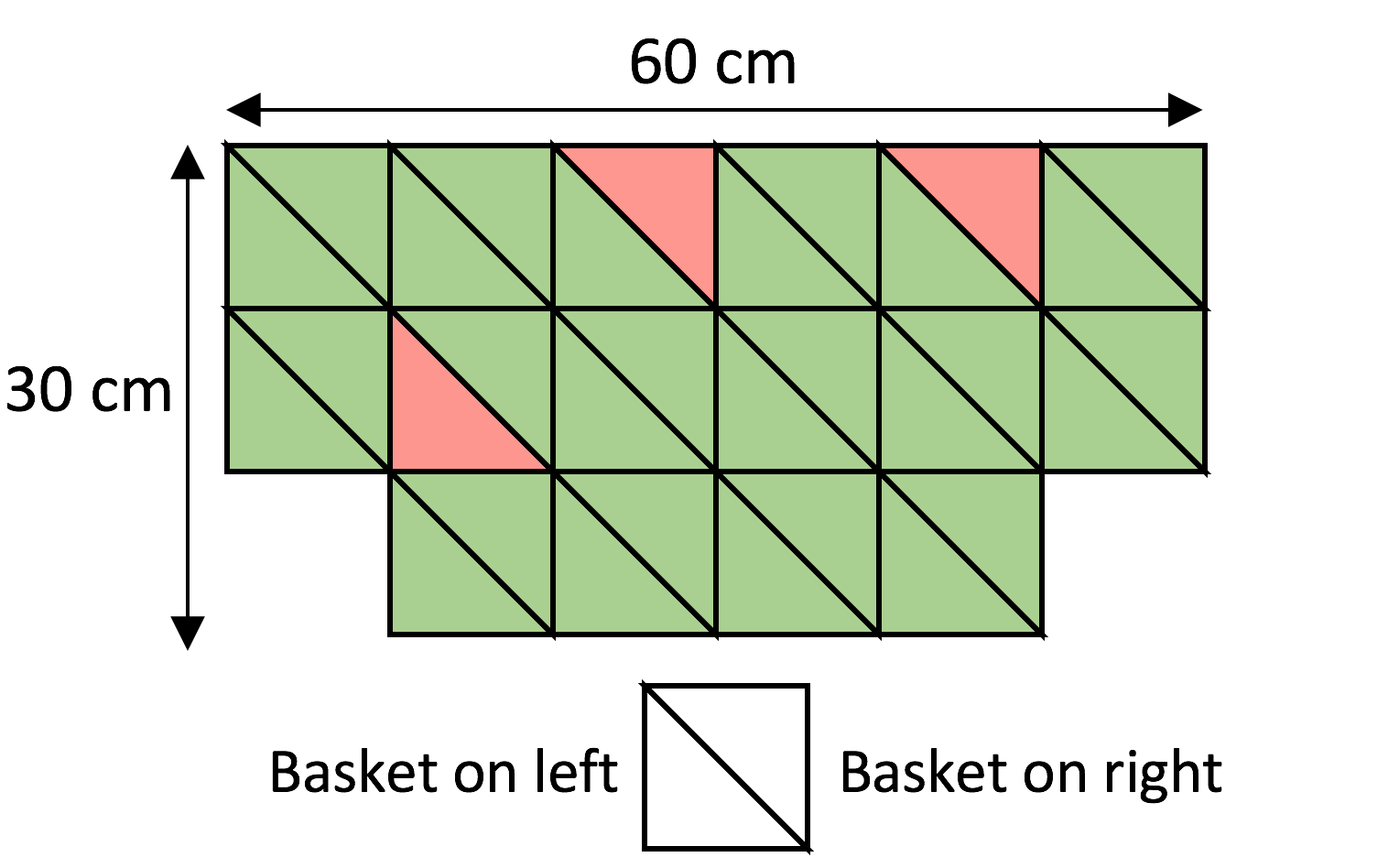}
\caption{Controller evaluation example. Green signifies a success, whilst red signifies a failure. In this case, success would be 91\%.}
\label{fig:how_test}
\end{wrapfigure} 

\begin{wrapfigure}{r}{7cm}
\centering
\includegraphics[width=0.9\linewidth]{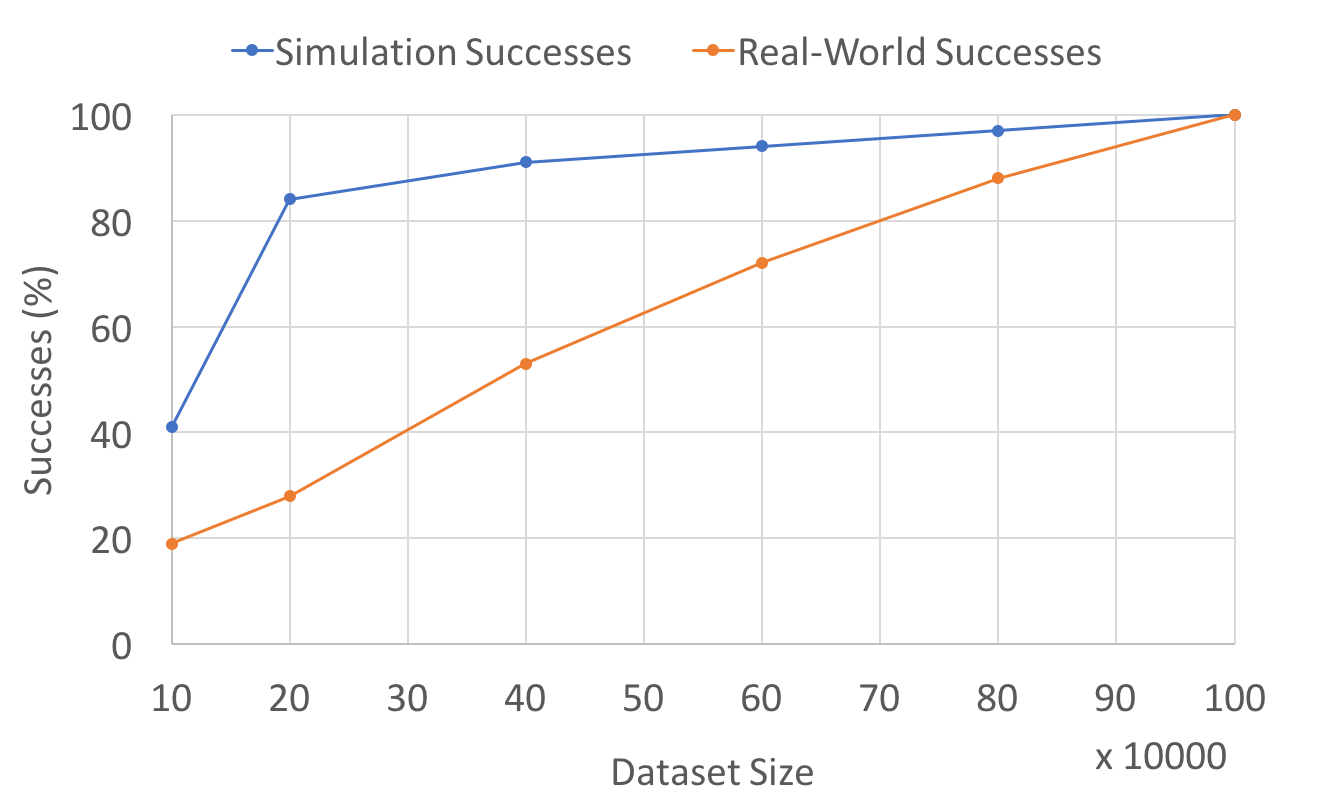}
\caption{How dataset size effects performance in both simulation and real world.}
\label{fig:graph_amount_of_data}
\end{wrapfigure}

\paragraph{Experimental Setup}
We first define how we evaluate the controller in both the simulation and real world. We place the cube using a grid-based method, where we split the area in which the cube was trained into a grid of $10cm \times 10cm$ cells. Using the grid, the cube can be in one of 16 positions, and for each position we run a trial twice with the basket on each side of the arm, resulting in 32 trials; therefore, all of our results are expressed as a percentage based on 32 trials. This number of trails is sufficient for a trend to emerge, as shown later in Figure \ref{fig:graph_amount_of_data}. In Figure \ref{fig:how_test}, we summarise the testing conditions, where each square represents a position, and the two triangles represent whether the basket was on the left or right side of the robot at that position.

\vspace{-8pt}

\paragraph{Altering Dataset Size}
To answer the first question of how the dataset size effects performance, we train several instances of the same network on dataset sizes ranging from 100,000 to 1 million images. Figure \ref{fig:graph_amount_of_data} shows the success rates in both simulation and real world, for the task in an environment with no distractors (such as in Figure \ref{fig:vrep_cam_variations}). The graph shows that a small dataset of 200,000 images achieves good performance in simulation, but 4 times more data is needed in order to achieve approximately the same success rate in the real world. Interestingly, both simulation and real world achieve 100\% for the first time on the same dataset size (1 million).

\vspace{-8pt}

\begin{figure}[]
\centering
\includegraphics[width=1.0\linewidth]{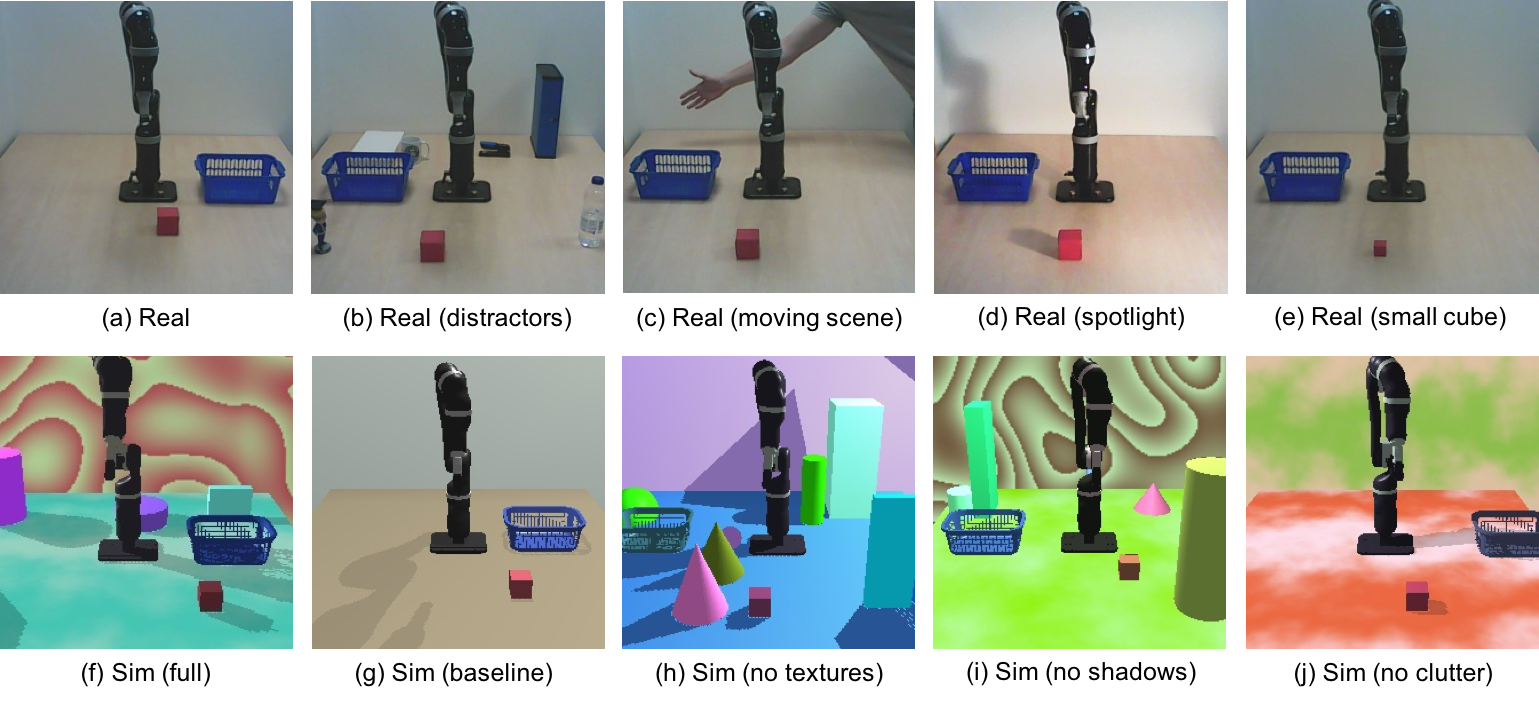}
\caption{Images (a) to (e) are taken from the real world, whilst images (f) to (j) are taken from simulation. The real world images show the testing scenarios from Table \ref{table:results}, whilst the simulated images shows samples from the training set from Table \ref{table:results}.}
\label{fig:comparison_images}
\vspace{-2.0mm}
\end{figure}

\paragraph{Robustness to New Environments}
We now turn our attention to Figure \ref{fig:comparison_images} and Table \ref{table:results}, which provide a summary of our results for the remaining research questions. First, we discuss the results in the top half of Table \ref{table:results}, where we evaluate how robust the network is to changes in the testing environment. Successes are categorised into \emph{cube vicinity} (whether the gripper's tip reached within $\approx 2cm$ of the cube), \emph{cube grasped} (whether the gripper lifted the cube off the table), and full task (whether the cube was reached, grasped, and dropped into the basket). The first two results show our solution achieving 100\% when tested in both simulation and the real world. Although the network was trained with distractors, it was not able to achieve 100\% with distractors in the real world. Note that the controller does not fail once the cube has been grasped, but rather fails during the reaching or grasping. The majority of failures in this case where when the cube was closest to the distractors in the scene. In the moving scene test, a person waved their arm back and forth such that each frame saw the arm in a different location. Reaching was not affected in this case, but grasping performance was. The moving camera test was performed by continuously raising and lowering the height of the tripod where the camera was mounted within a range of 2 inches during each episode. Although never experiencing a moving scene or camera motion during training, overall task success remained high at 81\% and 75\% respectively. To test invariance to lighting conditions, we aimed a bright spotlight at the scene and then moved the light as the robot moved. The results show that the arm was still able to recognise the cube, achieving a 84\% cube vicinity success rate, but accuracy in the grasp was affected, resulting in the cube only being grasped 56\% of the time. This was also the case when we replaced the cube with one that was half the length of the one seen in training. The 89\% vicinity success in comparison to the 41\% grasp success shows that this new object was too different to perform an accurate grasp, and the gripper would often only brush the cube. Interestingly, when the cube was replaced with a novel object such as a stapler or wallet, the task was occasionally successful. One explanation for this behaviour could be down to a clear colour discontinuity in comparison to the background in the area in which the cube is normally located.

\vspace{-8pt}

\paragraph{Ablation Study}
The bottom half of Table \ref{table:results} focuses on the final 3 questions regarding which aspects are important to randomise during simulation, and whether auxiliary tasks and joint angles improve performance. Firstly, we wish to set a baseline which shows that naively simulating the real world is difficult for getting high success in the real world. We generated a dataset based on a scene with colours close to the real world. The baseline is unable to succeed at the overall task, but performs well at reaching. This conclusion is in line with other work~\citep{james20163d, zhang2015towards}. It is clear that having no domain randomisation does not transfer well for tasks that require accurate control. Whilst observing the baseline in action, it would frequently select actions that drive the motors to force the gripper into the table upon reaching the cube. Training a network without distractors and testing without distractors yields 100\%, whilst testing with distractors unsurprisingly performs poorly. A common characteristic of this network is to make no attempt at reaching the cube, and instead head directly to above the basket. We tested our hypothesis that using complex textures yields better performance than using colours drawn from a uniform distribution. Although reaching is not affected, both grasping and full task completion degrade to 69\% and 44\% respectively. Moreover, swapping to the table illustrated in Figure \ref{fig:basket_move_timeline} leads to complete failure when using plain colours.

\begin{table}
  \centering
\begin{tabular}{ | c | c || c | c | c |  }
 \hline
 \multicolumn{2}{|c|}{Scenario} & \multicolumn{3}{|c|}{Successes} \\
 \hline
 Train & Test & Cube vicinity & Cube grasped & Full task\\
 \hline
 Sim (full)							& Sim (full)									& 100\% & 100\% & 100\%\\
 Sim (full) 							& Real										& 100\% & 100\% & 100\%\\
 Sim (full) 							& Real (distractors)							& 89\% & 75\% & 75\%\\
 Sim (full) 							& Real (moving scene)				& 100\% & 89\% & 89\%\\
 Sim (full) 							& Real (moving camera)				& 97\% & 89\% & 75\%\\
 Sim (full) 							& Real (spotlight)						& 84\% & 56\% & 56\%\\
 Sim (full) 							& Real (small cube)						& 89\% & 41\% & 41\%\\
 \hline
 Sim (baseline) 				& Real										& 72\% & 0\% & 0\%\\
 Sim (no distractors) 				& Real										& 100\% & 100\% & 100\%\\
 Sim (no distractors) 				& Real (distractors)					& 53\% & 0\% & 0\%\\
 Sim (no textures) 				& Real										& 100\% & 69\% & 44\%\\ 
 Sim (no moving cam) 		& Real										& 100\% & 9\% & 3\%\\
 Sim (no shadows) 				& Real										& 81\% & 46\% & 19\%\\
 Sim (no LSTM) 					& Real										& 100\% & 56\% & 0\%\\ 
 Sim (no auxiliary) 				& Real										& 100\% & 84\% & 84\%\\ 
 Sim (no joint angles) 		& Real										& 100\% & 44\% & 25\%\\

 \hline
\end{tabular}
\caption{Results based on 32 trials from a dataset size of 1 million images ($\approx 4000$ episodes) run on a square table in the real world. Successes are categorised into \emph{cube vicinity} (whether the arm reached within $\approx 2cm$ of the cube, based on human judgement), \emph{cube grasped}, and \emph{full task} (whether the cube was reached, grasped and dropped into the basket). The top half of the table focuses on testing the robustness of the full method, whilst the bottom half focuses on identifying what are the key aspects that contribute to transfer. A sample of the scenarios can be seen in Figure \ref{fig:comparison_images}.}
\label{table:results}
\vspace{-4.0mm}
\end{table}

Without moving the camera during training, the full task is not able to be completed. Despite this, target reaching seems to be unaffected. We cannot guarantee the position of the camera in the real world, but the error is small enough such that the network is able to reach the cube, but large enough that it lacks the ability to achieve a grasp. Results show that shadows play an important role following the grasping stage. Without shadows, the network can easily become confused by the shadow of both the cube and its arm. Once grasped, the arm frequently raises and then lowers the cube, possibly due to the network mistaking the shadow of the cube for the cube itself.

We now observe which aspects of the network architecture contribute to success when transferring the controllers. We alter the network in 3 distinct ways -- no LSTM, no auxiliary output, and no joint angles -- and evaluate how performance differs. Excluding the LSTM unit from the network causes the network to fail at the full task. Our reasoning for including recurrence in the architecture was to ensure that state was captured. As this is a multi-stage task, we felt it important for the network to know what stage of the task it was in, especially if it is unclear from the image whether the gripper is closed or not (which is often the case). Typical behaviour includes hovering above the cube, which then causes the arm to drift into unfamiliar states, or repeatedly attempting to close the gripper even after the cube has been grasped. Overall, the LSTM seems fundamental in this multi-stage task. The next component we analysed was the auxiliary outputs. The task was able to achieve good performance without the auxiliaries, but not as high as with the auxiliaries, showing the benefit of this auxiliary training. Finally, the last modification we tested was to exclude joint angles. We found that the joint angles helped significantly in keeping the gripper in the orientation that was seen during training. Excluding the joint angles often led to the arm reaching the cube vicinity, but failing to be precise enough to grasp. This could be because mistakes by the network are easier to notice in joint space than in image space, and so velocity corrections can be made more quickly before reaching the cube. 


\section{Conclusions}
\label{sec:conclusion}

In this work, we have shown transfer of end-to-end controllers from simulation to the real world, where images and joint angles are continuously mapped directly to motor velocities, through a deep neural network. The capabilities of our method are demonstrated by learning a long-horizon multi-stage task that is analogous to a simple tidying task, and involves locating a cube, reaching the cube, grasping the cube, locating a basket, and finally dropping the cube into the basket. We expect the method to work well for other multi-stage tasks, such as tidying other rigid objects, stacking a dishwasher, and retrieving items from shelves, where we are less concerned with dexterous manipulation. However, we expect that the method as is, would not work in instances where tasks cannot be easily split into stages, or when the objects require more complex grasps. Despite the simplicity of the method, we are able to achieve very promising results, and we are keen to explore the limits of this method in more complex tasks.

\clearpage

\acknowledgments{Research presented in this paper has been supported by
Dyson Technology Ltd. We thank the reviewers for their valuable feedback.}


\bibliography{corl.bib} 

\end{document}